%% file: main.tex
\documentclass{article}
\usepackage{iclr2027_conference,times}

\input{math_commands.tex}

\input{packages_commands}

\usepackage{hyperref}
\hypersetup{
  hidelinks,
  pdftitle={RAFT: Redundancy-Aware Input Filtering for Temporal Graph Neural Network Training},
  pdfauthor={Yue Dai, Sarah Zhou, Youtao Zhang}
}
\usepackage{url}

\title{RAFT: Redundancy-Aware Input Filtering for Temporal Graph Neural Network Training}

\author{Yue Dai \\
Department of Computer Science\\
Illinois Institute of Technology\\
Chicago, USA \\
\texttt{\{ydai21\}@illinoistech.edu} \\
\And
Sarah Zhou\\
BASIS Independent Silicon Valley\\
San Jose, CA, USA \\
\texttt{fire.and.ice7123@gmail.com} \\
\AND
Youtao Zhang \\
Department of Computer Science\\
University of Pittsburgh\\
Pittsburgh, USA \\
\texttt{youtao@pitt.edu}
}

\iclrfinalcopy
\begin{document}

\maketitle
\fancypagestyle{raftpreprint}{%
  \fancyhf{}%
  \fancyhead[L]{\small\textsc{A preprint}}%
  \fancyfoot[C]{\thepage}%
  \renewcommand{\headrulewidth}{0.4pt}%
  \renewcommand{\footrulewidth}{0pt}%
}
\pagestyle{raftpreprint}
\thispagestyle{raftpreprint}
\begin{abstract}
   \input{sections/abstract} 
\end{abstract}
\section{Introduction}
\input{sections/introduction}
\section{Background}
\label{sec:background}
\input{sections/background}
\section{Motivation: which temporal inputs are necessary?}
\label{sec:diagnostic}
\input{sections/motivation}
\section{RAFT design}
\label{sec:method}
\input{sections/method}
\section{Evaluation}
\label{sec:eval}
\input{sections/evaluation}
\section{Conclusion}
\input{sections/conclusion}
\bibliography{iclr2027_conference}
\bibliographystyle{iclr2027_conference}
\clearpage
\appendix
\input{sections/appendix}

\end{document}

%% file: math_commands.tex
\usepackage{amsmath,amsfonts,bm}

\def\eqref#1{equation~\ref{#1}}

\def\1{\bm{1}}

\DeclareMathAlphabet{\mathsfit}{\encodingdefault}{\sfdefault}{m}{sl}
\SetMathAlphabet{\mathsfit}{bold}{\encodingdefault}{\sfdefault}{bx}{n}



%% file: packages_commands.tex
\usepackage{tikz}
\usepackage[ruled,vlined,noend]{algorithm2e}
\usepackage{soul}
\usepackage{adjustbox}
\usepackage{empheq}
\usepackage{amsmath,amssymb,graphicx,booktabs,array,multirow}

\newcommand{\squishlist}{
   \begin{list}{$\bullet$}
    { 
    \setlength{\itemsep}{0pt}      \setlength{\parsep}{0pt}
      \setlength{\topsep}{3pt}       \setlength{\partopsep}{0pt}
      \setlength{\listparindent}{-2pt}
      \setlength{\itemindent}{-5pt}
      \setlength{\leftmargin}{1em} \setlength{\labelwidth}{0em}
      \setlength{\labelsep}{0.5em} } }

\newcommand{\squishend}{
    \end{list}  }

\usepackage{xcolor}
\definecolor{myblue}{RGB}{0,102,204}

\newcommand{\OurMethod}{RAFT}

%% file: sections/abstract.tex
Temporal graph neural networks (TGNNs) have shown strong performance on dynamic graph learning tasks by continuously updating node memories from evolving interactions (i.e., events) and leveraging these memories for downstream predictions. As graphs grow rapidly in modern applications, training requires processing an increasing number of events; meanwhile, each event can update node memories used by later events, creating dependencies that constrain parallel execution and limit the benefits of additional computing resources. These costs motivate an upstream question: \emph{are all events, and the dependencies they introduce, necessary for learning?}
We investigate this question through two roles of temporal events: prediction targets and historical events sampled as inputs. We observe that targets involving stable node memories tend to exhibit lower prediction loss, while filtering historical events associated with stable memories can retain similar training-loss trajectories in several evaluated settings. Together, these observations suggest an opportunity to identify input redundancy through changes in node memories.
Building on this opportunity, we propose \OurMethod{}, a redundancy-aware input-filtering framework for TGNN training. \OurMethod{} tracks memory changes during training and uses them to guide two filters: a root-event filter removes selected prediction targets before training, and a support-event filter excludes selected historical context during sampling. Filtering inputs before their downstream preparation and processing reduces the work they trigger.
Across three TGNN models and five datasets, \OurMethod{} reduces training inputs by 35.23\% on average and up to 97.11\%, while reducing recorded training time by up to 52.56\% with negligible changes to model precisions.

%% file: sections/introduction.tex
Dynamic graphs are central to applications such as recommendation, social interaction analysis, and temporal knowledge reasoning~\citep{trivedi2019dyrep,kumar2019predicting,rossi2020temporal,zhang2023tiger,you2022roland,wang2021apan,zhou2022tgl,kazemi2020representation,xu2020inductive,sankar2020dysat,wang2024tglite}. In these applications, relationships evolve over time, requiring learning models to capture both structural patterns and temporal dependencies. Temporal graph neural networks (TGNNs) address this need by learning from evolving events, where each event describes a temporal interaction between a source node and a destination node, such as adding an edge or a node to existing graphs~\citep{kumar2018community,kumar2019predicting,rossi2020temporal,belli2020privacy,chen2019lstm,jia2018tequila,chaintreau2007impact}. In particular, recent TGNNs maintain per-node states---often called node memories---that summarize historical interactions that occurred around these nodes~\citep{kumar2019predicting,rossi2020temporal,wang2021apan,zhou2022tgl,dai2025cascade,zhou2023disttgl,dai2025memfreezing}. Each incoming event can update the memories of its participating nodes, which then support subsequent predictions. These memory-based designs allow TGNNs to capture changing interaction histories and achieve strong performance on dynamic graph learning tasks.

As graphs grow rapidly in modern applications, TGNN training must process an increasing number of events. However, scaling TGNN training is challenging because events cannot always be processed independently.
An earlier event may update a node memory needed by a later prediction; for example, successive interactions involving the same user may require the later prediction to incorporate the earlier memory update. These dependencies constrain parallel execution and limit the benefits of additional computing resources~\citep{dai2025cascade,gao2024etc}. Meanwhile, processing events requires fetching node and edge features and moving data between CPUs and GPUs, adding substantial preparation costs. 
Existing systems address these challenges through dependency-aware batching and scheduling~\citep{dai2025cascade,gao2024etc}, efficient data preparation and execution~\citep{zhou2022tgl,wang2024tglite}, and distributed training~\citep{zhou2023disttgl}. Yet their efficiency is still significantly bounded by the natural dependency of input events. If processing an event introduces both preparation costs and memory-update dependencies, determining whether that event needs to participate in training could address these costs at their source. We therefore raise a fundamental question: 
\begin{quote}
\textit{Are all events from the training dynamic graphs, and the dependencies they introduce, necessary for learning?}  
\end{quote}

To answer this question, we examine two roles of an event, a temporal interaction between a source node and a destination node. A \textit{root event} is a prediction target that contributes to the training loss, whereas a \textit{support event} is a historical interaction sampled as input for that prediction. The same interaction can serve both roles at different times. At the root-event level, we observe that targets whose endpoint memories are \textit{stable} (i.e., changing little during updates) tend to have lower prediction loss, suggesting that the model already predicts these targets well. At the support-event level, we observe that omitting historical interactions associated with stable memories can retain similar training-loss trajectories, suggesting that some historical context provides limited additional benefit. Together, these observations reveal an opportunity to use memory stability to identify redundancy in both prediction targets and their historical inputs. Reducing unnecessary inputs at these two levels could avoid the data preparation and computation they would otherwise trigger.

Building on this opportunity, we propose \OurMethod{}, a redundancy-aware input-filtering framework for efficient TGNN training. \OurMethod{} tracks memory changes during executed updates and uses the resulting stability information to guide two filters. First, a root-event filter removes selected root events whose source and destination memories are stable, before these targets enter subsequent training stages. Second, a support-event filter excludes selected support events during temporal sampling, reducing the historical context returned for retained targets. Both filters use information from already executed memory updates, avoiding an additional prediction pass merely to decide whether an input should be filtered. By making these decisions before downstream feature preparation and model computation, \OurMethod{} reduces the work triggered by selected training inputs.
Our evaluation across three TGNN models and five datasets shows that \OurMethod{} reduces training inputs by 35.23\% on average and up to 97.11\%. The training time decreases by 17.80\% on average and up to 52.56\%, with mean absolute relative changes in validation AP and AUC of 0.13\% and 0.14\%, respectively.
We summarize our contributions as follows:
\begin{itemize}
    \item We investigate input redundancy in memory-based TGNN training and identify two filtering opportunities through observations of prediction targets and sampled historical inputs.

    \item We propose \OurMethod{}, a redundancy-aware input-filtering framework that uses shared memory-stability information to select prediction targets before training and historical context during sampling, reducing downstream preparation and computation.

    \item We evaluate \OurMethod{} across three TGNN models and five datasets, demonstrating substantial input reduction and training-time savings while assessing the associated changes in prediction quality.
\end{itemize}

%% file: sections/background.tex
\subsection{Dynamic graphs and memory-based TGNNs}

Dynamic graphs describe interactions that evolve over time. They are commonly represented as either \textit{discrete-time dynamic graphs} (DTDGs), which organize graph evolution into snapshots, or \textit{continuous-time dynamic graphs} (CTDGs), which represent it as an ordered stream of timestamped events~\citep{trivedi2019dyrep,kumar2019predicting,rossi2020temporal,zhang2023tiger,you2022roland,wang2021apan,zhou2022tgl,kazemi2020representation,xu2020inductive,sankar2020dysat,wang2024tglite}. By preserving the timing and ordering of individual interactions, CTDGs capture fine-grained temporal information and naturally represent asynchronous graph evolution. We focus on memory-based TGNNs for this setting.

Formally, we represent a CTDG as an event stream:
\begin{equation}
    \mathcal E = \{e_i=(u_i,v_i,t_i,\mathbf x_i)\}_{i=1}^{N},
\end{equation}
where $u_i$ and $v_i$ are the interacting nodes, $t_i$ is the timestamp, and $\mathbf x_i$ contains the available interaction features. For example, a purchase connects a user to an item at a particular time. Temporal link prediction uses the preceding interaction history to predict whether two nodes will interact.

To encode this evolving history, a memory-based TGNN maintains a vector $\mathbf m_v$ for each node $v$~\citep{kumar2019predicting,rossi2020temporal,wang2021apan,dai2025cascade,gao2024etc,dai2025memfreezing,zhou2022tgl}. As events are processed, their information is incorporated into node memories and carried forward to subsequent predictions. For prediction, the model computes an \emph{embedding}, a vector representation derived from the available node memory and, in models with temporal neighborhood aggregation, sampled historical interactions. These embeddings are then used to score candidate interactions. Node memories therefore provide the evolving historical context on which predictions depend.

\subsection{TGNN training and memory dependencies}

TGNN training typically processes events in chronological mini-batches. The prediction targets in the current batch $B$ are called \textit{root events}. For models that use temporal neighborhoods, a sampler retrieves earlier interactions $\mathcal S(B)$ as inputs for these predictions; these historical inputs are called \textit{support events}. In the purchase example, the purchase being predicted is a root event, while sampled earlier purchases provide support. The same interaction can serve as a root when it is processed and as support for later predictions.

\begin{figure}[t]
    \centering
    \includegraphics[width=\linewidth]{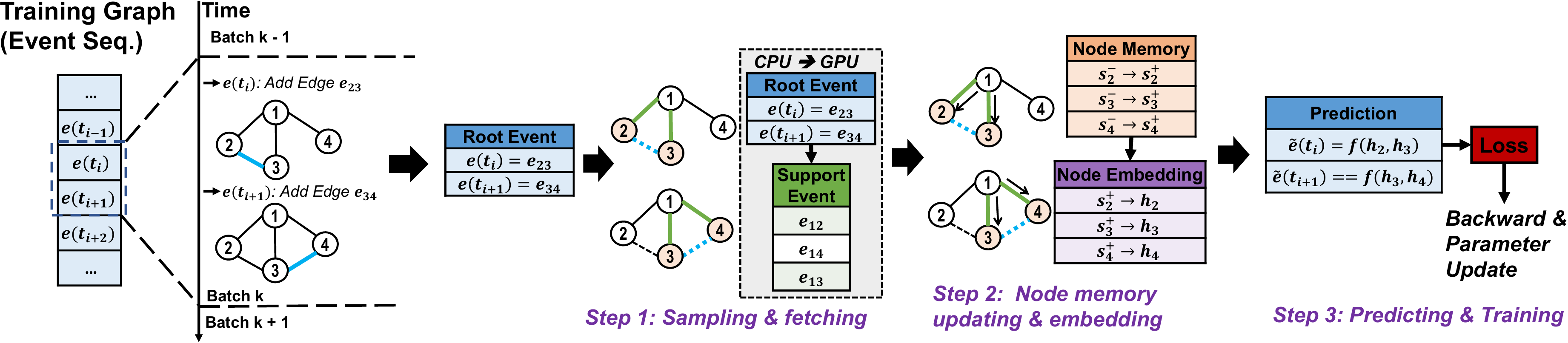}
    \caption{Batched memory-based TGNN training. Root events (e.g., $e_{23}$ and $e_{34}$) are prediction targets, while support events (e.g., $e_{12}$, $e_{13}$, and $e_{14}$) provide historical inputs. Step~1 samples support events and fetches the required information. Step~2 updates node memories ($s_v^{-}\rightarrow s_v^{+}$) and computes embeddings $h_v$. Step~3 predicts root events (e.g., $\tilde e(t_i)=f(h_2,h_3)$), computes the loss, and updates model parameters through backpropagation.}
    \label{fig:background}
\end{figure}

Figure~\ref{fig:background} illustrates the training pipeline. The trainer first samples support events and fetches the required node and edge features. The model uses the available memories and sampled context to compute embeddings and predict the root events. Training compares scores for observed interactions with those for sampled noninteractions, computes the loss, and updates model parameters through backpropagation. Memory updates carry information from processed interactions forward to subsequent predictions. 
Historical interactions may be sampled repeatedly for different targets and across epochs; we count each occurrence as a separate event when measuring volume of historical inputs.

These memory operations impose a fundamental constraint on training parallelism. If a later prediction requires a memory produced by an earlier interaction, their memory-dependent operations cannot execute independently. Such dependencies restrict which events can be grouped or executed concurrently, so enlarging batches or adding computing resources does not automatically yield proportional throughput gains. Meanwhile, both root and support events incur feature preparation and computation costs.

Existing systems seek greater parallelism through dependency-aware batching and scheduling, grouping compatible events or selectively relaxing memory dependencies~\citep{dai2025cascade,gao2024etc}. Other frameworks improve sampling, data preparation, and distributed execution~\citep{zhou2022tgl,wang2024tglite,zhou2023disttgl}. These approaches improve how events are executed. We examine an earlier decision: whether each event needs to generate that work in the first place. Determining which prediction targets and historical inputs are necessary offers an opportunity to avoid their associated preparation and computation before scheduling and execution. This motivates our investigation of input redundancy as a way to reduce training costs at their source.

%% file: sections/motivation.tex
Since memory updates create dependencies between events, we investigate whether changes in node memories can reveal redundancy in training inputs. 

\begin{figure}[t]
  \centering
  \includegraphics[width=\linewidth]{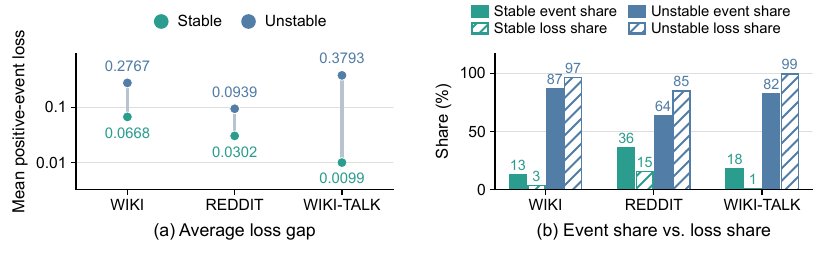}
  \vspace{-0.35in}
  \caption{Root-event loss during TGN training on WIKI, REDDIT, and WIKI-TALK. (a) Stable root events have lower mean positive-event loss than unstable events. (b) Their share of training loss is considerably smaller than their share of root-event counts.}
  \label{fig:root}
\end{figure}

\paragraph{Redundancy in root events}
Root events provide supervision, but targets that the model already predicts well contribute less to the training loss. 
To quantify this effect, we evaluate the contribution of each root event to the training loss by training TGN~\citep{rossi2020temporal} on the Wikipedia (WIKI), Reddit (REDDIT)~\citep{kumar2019predicting}, and Wikipedia-talk(WIKI-TALK)~\citep{paranjape2017motifs} datasets.
At first, we divide the root events into \textit{stable} and \textit{unstable} sets based on the stability of their source and destination nodes. 
We classify a root event as \emph{stable} when, for each of its endpoints, the cosine similarity between the node's memories before and after its most recent update exceeds 0.9; otherwise, the event is \emph{unstable}.
Figure~\ref{fig:root}(a) shows that stable root events have lower average training loss on all three datasets. Their share of training loss is also smaller than their share of root-event counts, as shown in Figure~\ref{fig:root}(b). For example, on WIKI-TALK, stable events account for 17.69\% of root events but only 0.56\% of training loss. These results suggest that memory stability can identify targets the model already predicts well, creating an opportunity to filter selected roots before they trigger historical sampling, feature preparation, and training.

\begin{figure}[t]
  \centering
  \includegraphics[width=\linewidth]{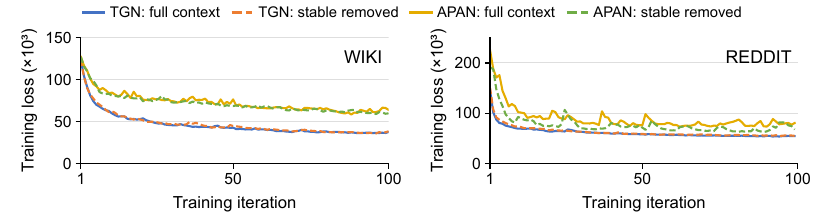}
  \vspace{-0.35in}
  \caption{Training-loss curves for TGN and APAN on WIKI and REDDIT with and without stable support events. Solid curves show training with all sampled support events; dashed curves show training with support events whose source nodes are stable removed.}
  \label{fig:support-motivation}
\end{figure}

\paragraph{Redundancy in sampled support events}
Even when a root event is retained, its prediction may not require all of the sampled history. Node memories already summarize previous interactions, suggesting that some historical inputs repeat information the model has captured. We examine this opportunity by removing support events whose source nodes are stable while retaining the prediction targets.
To quantify this effect, we evaluate the impact of support events on training convergence in TGN~\citep{rossi2020temporal} and APAN~\citep{wang2021apan} on the WIKI and REDDIT datasets~\citep{kumar2019predicting}. 
Similar to the root-event analysis, we first divide sampled support events into two categories based on the stability of their source nodes: \textit{stable support events}, whose source nodes undergo only minor memory changes in their most recent updates, and \textit{unstable support events}, whose source nodes remain actively changing. 
Intuitively, unstable support events are more likely to inject fresh temporal signals from neighboring nodes, whereas stable support events mainly provide repeated information. 
We then compare TGNN training with full support events and with stable support events removed. 
As shown in Figure~\ref{fig:support-motivation}, even with these events removed, TGN's training-loss curves remain close on both datasets. APAN follows a similar declining trajectory on WIKI and generally exhibits lower training loss under filtering on REDDIT. These observations suggest an opportunity to reduce sampled support events while maintaining similar loss-reduction trends.

Together, these observations motivate using memory stability to select root events before training and support events during sampling. Appendix~\ref{app:extra-motivation} extends the analysis to more models and datasets.

%% file: sections/method.tex

\subsection{Overview}
Building on these opportunities, we propose \OurMethod{}, a redundancy-aware input-filtering framework for TGNN training. Guided by node memory changes, \OurMethod{} reduces training inputs at two levels. The root-event filter removes stable prediction targets, which tend to have low prediction loss, before they enter downstream training stages. The support-event filter prunes sampled historical inputs associated with stable source nodes, reducing the context processed for retained targets. An adaptive controller maintains node stability information and uses training-loss feedback to determine when these filters operate. Figure~\ref{fig:pipeline} illustrates the workflow with a small input graph.

\subsection{Root-event Filter}
Stable root events tend to have smaller prediction losses, suggesting
an opportunity to remove targets the model already predicts well.
To identify these events, \OurMethod{} compares each node's
memories before and after its latest executed update using
cosine similarity:
\begin{equation}
    c_v = \operatorname{cos}
    \left(\mathbf m_v^{-}, \mathbf m_v^{+}\right),
    \label{eq:memory-stability}
\end{equation}
where $\mathbf m_v^{-}$ and $\mathbf m_v^{+}$ are node $v$'s memories
before and after that update. A node is marked stable when $c_v>\tau$, and its stability flag
is refreshed after each executed memory update. We use
$\tau=0.9$ by default, consistent with the motivation study in
Section~\ref{sec:diagnostic}, and evaluate sensitivity to this
threshold in Section~\ref{subsec:eval:sens}.

Given a root batch $B$, the enabled filter removes events whose
source and destination nodes are both stable:
\begin{equation}
    \widetilde B
    =
    B \setminus
    \left\{
        e=(u,v,t,\mathbf x)\in B
        \;\middle|\;
        c_u>\tau \;\land\; c_v>\tau
    \right\}.
    \label{eq:root-filter}
\end{equation}
The resulting batch $\widetilde B$ contains the retained prediction
targets in their original temporal order. Each filtering decision
requires only two stored stability-flag lookups.

As shown in Figure~\ref{fig:pipeline}, the original root events are
$A\!\to\!B$ and $C\!\to\!D$. The filter retains $A\!\to\!B$ because
$A$ has a changing memory, even though $B$ is stable. It removes
$C\!\to\!D$ because both endpoints are stable. Removing a root before
downstream processing avoids the historical sampling, input
preparation, prediction, and memory-update work that its execution
would trigger, reducing the inputs entering the dependency-constrained
training pipeline.

\subsection{Support-event Filter}
Root filtering reduces the number of prediction targets, but each retained target may still retrieve redundant historical inputs. The support-event filter reduces this context by checking the source-node stability of each sampled historical event.

Let $S(\widetilde B)$ denote the collection of sampled support-event
instances for the retained roots. When enabled, the filter removes
instances whose source nodes are stable:
\begin{equation}
    \widetilde S
    =
    S(\widetilde B) \setminus
    \left\{
        e_j=(u_j,v_j,t_j,\mathbf x_j)\in S(\widetilde B)
        \;\middle|\;
        c_{u_j}>\tau
    \right\}.
    \label{eq:support-filter}
\end{equation}
Here, $j$ indexes sampled occurrences. Each decision requires one
source-node stability-flag lookup. Filtering reduces the sampled
inputs supplied to training while leaving the underlying interaction
history available for future sampling.

\begin{figure}[t]
  \centering
  \includegraphics[width=\linewidth]{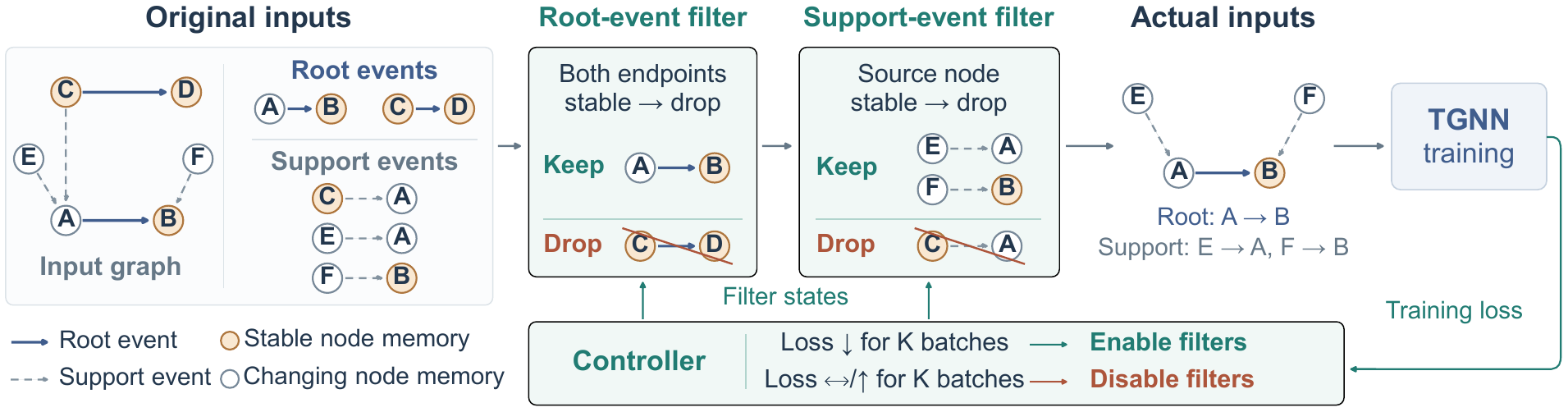}
\caption{Overview of \OurMethod{}. Root filtering removes targets with two stable endpoints, while support filtering removes sampled historical inputs with stable source nodes. The retained inputs enter TGNN training, whose loss feeds back to the controller. Each loss-trend condition requires $K$ consecutive batches; otherwise, the controller retains its current state.}
  \label{fig:pipeline}
\end{figure}

As shown in Figure~\ref{fig:pipeline}, the retained root $A\!\to\!B$ samples
support events $E\!\to\!A$, $F\!\to\!B$, and $C\!\to\!A$.
The filter retains the first two because $E$ and $F$ have changing
memories. It removes $C\!\to\!A$ because source $C$ is stable,
regardless of destination $A$'s changing memory. The resulting input
preserves the prediction target $A\!\to\!B$ while reducing the
historical context passed to downstream preparation and computation.

\subsection{Adaptive Filter Controller}
The controller initially enables the filters applicable to the model;
support filtering applies only when the model samples historical
interactions. It then compares each executed batch's loss with that
of the preceding executed batch. While filtering is enabled,
$K$ consecutive batches without a loss decrease disable the filters,
where unchanged and increased loss both count as non-decreases.
While filtering is disabled, $K$ consecutive batches with decreasing
loss re-enable them. 
We empirically set $K=5$. 
An opposite trend resets the corresponding
consecutive count, and each state transition resets the counts.
Otherwise, the controller retains its current state. Disabled filters
pass their inputs through unchanged, while normal memory updates
continue to refresh node stability flags.
 

%% file: sections/evaluation.tex
\OurMethod{} addresses training efficiency by selecting the events that enter the training pipeline. Its benefit therefore depends on how much input can be omitted, what this changes for learning, and how the reduction translates into saved work. We organize the evaluation around four research questions:
\begin{enumerate}
  \item[\textbf{RQ1.}] How much training input can \OurMethod{} remove, and how does this affect prediction quality?
  \item[\textbf{RQ2.}] Do fewer training inputs translate into shorter training time?
  \item[\textbf{RQ3.}] How much does each filter contribute, individually and together?
  \item[\textbf{RQ4.}] How sensitive are input reduction and prediction quality to the memory-stability threshold?
\end{enumerate}

\subsection{Experimental Setup}
\textbf{Benchmark and Testbed.} We evaluate TGN~\citep{rossi2020temporal}, APAN~\citep{wang2021apan}, and JODIE~\citep{kumar2019predicting} on five datasets: WIKI and REDDIT~\citep{kumar2019predicting}, WIKI-TALK and STACK-OVERFLOW (abbreviated as SX-FULL)~\citep{paranjape2017motifs}, and GDELT~\citep{leetaru2013gdelt,zhou2022tgl}. Our primary comparison adds \OurMethod{} to a state-of-the-art efficient training framework, Cascade~\citep{dai2025cascade} while retaining the underlying model; we also evaluate integration with TGL~\citep{zhou2022tgl}. Experiments use an AMD EPYC 7742 64-core CPU and an NVIDIA A100 40\,GB GPU with CUDA 11.6.

\textbf{Input reduction.}
We collectively refer to root events and support events as \emph{training inputs}: root events provide prediction targets, while support events provide historical context for those targets. Input reduction measures the percentage decrease in the corresponding event count relative to the unfiltered baseline, calculated as $100(1-N_{\mathrm{RAFT}}/N_{\mathrm{base}})\%$. Depending on the experiment, $N$ counts prediction targets or sampled historical-event occurrences. An interaction sampled multiple times contributes multiple occurrences. 

\textbf{Prediction quality and training time.}
We report validation average precision (AP) and area under the ROC curve (AUC). Quality changes are expressed as percentages relative to the corresponding baseline.
\emph{Training-time reduction} is $100(1-T_{\mathrm{RAFT}}/T_{\mathrm{base}})\%$, where both times use the corresponding framework's recorded training time. This field subtracts the separately timed filtering and updater regions from the accumulated training interval; Appendix~\ref{app:timer-definitions} defines its scope and the profiling stages. Positive reductions indicate shorter recorded training time. Reported averages are arithmetic means across the stated configurations. Appendix~\ref{app:setup} provides the dataset statistics, training configurations, evaluation protocol, and input-counter definitions.

\subsection{Input Reduction, Prediction Quality, and Efficiency (RQ1--RQ2)}
\label{sec:inputs}

As shown in Table~\ref{tab:main}, across three TGNN models and five datasets, \OurMethod{} reduces training inputs by \textbf{35.23\%} on average and up to \textbf{97.11\%}. The benefits are particularly pronounced on large event streams. For TGN, input reduction increases from 16.74\% on WIKI, which contains approximately 157 thousand events, to 78.66\% on SX-FULL and 93.13\% on GDELT, which contain 63.50 million and 191.29 million events, respectively. TGN therefore processes only 21.34\% and 6.87\% of the baseline's sampled inputs on these two large graphs. APAN also achieves a substantially larger reduction on SX-FULL than on WIKI, 46.31\% versus 6.55\%. JODIE contributes through root-event filtering and reaches the largest reduction in the table, 97.11\%, on GDELT. These results highlight the value of selecting inputs at their source: on large event streams, omitting redundant inputs can avoid substantial downstream work.

Prediction quality remains close to the baseline across the main comparisons. Across the all comparisons, the average relative changes in validation AP and AUC are \textbf{0.13\%} and \textbf{0.14\%}, respectively. On TGN/SX-FULL, the 78.66\% input reduction accompanies relative AP and AUC decreases of only 0.07\% and 0.09\%. TGN/GDELT reduces sampled inputs by 93.13\%, with relative AP and AUC decreases of 0.59\% and 0.45\%, respectively. These large-graph comparisons show that substantial input reduction can coexist with prediction quality close to the baseline.

\begin{table}[t]
  \centering\small
  \caption{Comparison between the original models trained under Cascade (Base) and the same models trained with \OurMethod{}. Input reduction indicates the percentage reduction in input events. AP and AUC are validation metrics, where higher values indicate better prediction quality. Time reports the reduction (\%) in recorded training time relative to the Base models. Parenthesized percentages report relative changes in AP and AUC from Base; arrows indicate increases or decreases.}
  \label{tab:main}
  \setlength{\tabcolsep}{4pt}
  \input{tables/eval_main}
\end{table}

The training-time results show that input selection can deliver substantial practical savings on large graphs. Across all comparisons with Cascade, recorded training time decreases by \textbf{17.80\%} on average and up to \textbf{52.56\%}. For TGN, the reduction grows from 5.69\% on WIKI to 40.88\% on SX-FULL and 52.56\% on GDELT. APAN also benefits more on SX-FULL, where its time reduction reaches 27.47\%, compared with 3.91\% on WIKI. Among TGN's five datasets, the two largest event streams yield its largest input and training-time reductions, illustrating the benefit of avoiding redundant work as event-processing workloads grow. Similarly, integrating \OurMethod{} with TGL reduces training time by \textbf{12.42\%} on average. Appendix~\ref{app:timing} provides the complete TGL comparisons and an efficiency analysis of the underlying training stages.

\subsection{Contribution of the Two Filters (RQ3)}
\label{sec:ablation}
To study the contribution of the two filters, we evaluate their individual and combined performance and compare them with the baseline models. Root filtering removes selected prediction targets and avoids their sampling requests, while support filtering reduces the historical context supplied to retained targets. Comparing Base, Root only, Support only, and Both reveals how these two selection points contribute to training efficiency.

As shown in Table~\ref{tab:ablation}, combining the filters achieves the largest sampled-input reduction in all four settings. For TGN/REDDIT, root-only and support-only filtering remove 50.42\% and 56.02\% of sampled inputs, respectively, while their combination reaches 67.68\%. The larger contributor varies across workloads: root filtering removes more input on TGN/WIKI and both APAN settings, whereas support filtering contributes more on TGN/REDDIT. Applying both filters captures these complementary opportunities without relying on either selection point alone.

The additional input reduction also brings greater time savings in several settings. For TGN/REDDIT, combined filtering reduces recorded training time by 48.32\%, compared with 36.90\% for root-only and 37.69\% for support-only filtering, with a relative AUC decrease of only 0.03\%. On APAN/REDDIT, combined filtering reduces recorded training time by 9.60\% while increasing AUC by 0.02\%. All four combined configurations reduce recorded training time; their largest relative AUC decrease is 0.49\%, on APAN/WIKI. 

\subsection{Sensitivity to the Stability Threshold (RQ4)}
\label{subsec:eval:sens}
We further examine how the memory-stability threshold $\tau$ affects input reduction and prediction quality by evaluating $\tau\in\{0.8,0.9,0.95\}$ with both filters enabled for 100 epochs (Table~\ref{tab:sensitivity-main}). For a given memory state, lowering $\tau$ classifies more nodes as stable and makes more events eligible for filtering. Each model--dataset sweep uses the same unfiltered baseline and holds the remaining settings fixed.

The input-reduction benefit persists across all three thresholds. On TGN/REDDIT, \OurMethod{} removes 51.62--52.98\% of sampled inputs, with relative validation AUC decreases below 0.02\%. APAN/REDDIT also achieves substantial reductions of 22.97--27.53\%, with a maximum relative AUC increase of approximately 0.02\%. On WIKI, the reductions remain consistent across thresholds, ranging from 13.39\% to 14.10\% for TGN and from 6.33\% to 6.80\% for APAN. Thus, the opportunity to remove historical inputs is present across a range of stability thresholds.

Prediction quality helps distinguish operating points with similar input reductions. On APAN/WIKI, $\tau=0.9$ removes 6.80\% of inputs and increases AUC by 0.10\% relative to the baseline, whereas $\tau=0.95$ removes 6.33\% with a relative AUC decrease of 0.37\%. A stricter threshold therefore does not necessarily provide a better reduction--quality trade-off. Filtering can change subsequent inputs and memory updates, so cumulative reductions need not vary monotonically with the threshold. These results motivate selecting $\tau$ according to the observed prediction quality while retaining the input-reduction benefit. Appendix~\ref{app:study-settings} provides the experimental settings.

\begin{table}[t]
  \centering\small
  \caption{Comparison under different filter configurations. Base disables both filters; Root only and Support only enable the corresponding filter; Both enables both. Input and Time report percentage reductions in sampled inputs and training time relative to Base. AUC is validation AUC; parenthesized percentages and arrows show its relative change. Each model--dataset study uses a shared baseline.}
  \label{tab:ablation}
  \setlength{\tabcolsep}{3.8pt}
  \input{tables/eval_components}
\end{table}

\begin{table}[t]
  \centering\small
  \caption{Sensitivity of \OurMethod{} to the memory-stability threshold $\tau$, with both filters enabled for 100 epochs. Input reports sampled-input reduction (\%); $\Delta$AUC reports the relative change in validation AUC (\%). Within each model--dataset study, all thresholds share the same unfiltered baseline. Positive and negative $\Delta$AUC values indicate increased and decreased AUC, respectively.}
  \label{tab:sensitivity-main}
  \input{tables/eval_sensitivity_main}
\end{table}

\paragraph{Discussion.}
The results highlight three implications for efficient TGNN training. First, large interaction histories offer substantial opportunities for input reduction. TGN achieves its largest input and training-time reductions on SX-FULL and GDELT, showing the practical value of redundancy-aware selection on large graphs. Second, redundancy exists in both supervision and historical context: combining root and support filtering removes more inputs than either filter alone in all four component settings. Selecting targets and selecting their context therefore provide complementary benefits.
Third, input selection can complement existing training optimizations. The additional training-time reductions when integrating \OurMethod{} with Cascade~\citep{dai2025cascade} and TGL~\citep{zhou2022tgl} show that redundancy remains after improvements to batching and execution. Together, these findings motivate jointly optimizing which events participate in training and how the retained events are processed. Future work could combine memory-guided input selection with dependency-aware batch formation and explore its benefits in distributed training.

%% file: tables/eval_main.tex
\begingroup
\small
\setlength{\tabcolsep}{2pt}
\begin{tabular*}{\linewidth}{@{\extracolsep{\fill}}lccccccc@{}}
\toprule
\multirow{2}{*}{Dataset} & \multirow{2}{*}{Model} & \multirow{2}{*}{\shortstack{Input\\Reduction ($\downarrow$\%)}} & \multicolumn{2}{c}{Average Precision(AP)} & \multicolumn{2}{c}{AUC} & \multirow{2}{*}{\shortstack{Time\\Reduction ($\downarrow$\%)}} \\
\cmidrule(lr){4-5}\cmidrule(lr){6-7}
& & & Base & \OurMethod{} & Base & \OurMethod{} & \\
\midrule
\multirow{3}{*}{WIKI} & TGN & 16.74 & 0.9945 & 0.9947\,{\footnotesize $(0.02\%\uparrow)$} & 0.9946 & 0.9948\,{\footnotesize $(0.01\%\uparrow)$} & 5.69 \\
 & APAN & 6.55 & 0.9751 & 0.9779\,{\footnotesize $(0.28\%\uparrow)$} & 0.9787 & 0.9805\,{\footnotesize $(0.18\%\uparrow)$} & 3.91 \\
 & JODIE & 2.02 & 0.9832 & 0.9840\,{\footnotesize $(0.09\%\uparrow)$} & 0.9845 & 0.9857\,{\footnotesize $(0.12\%\uparrow)$} & 5.77 \\
\midrule
\multirow{3}{*}{REDDIT} & TGN & 67.68 & 0.9983 & 0.9980\,{\footnotesize $(0.03\%\downarrow)$} & 0.9984 & 0.9982\,{\footnotesize $(0.02\%\downarrow)$} & 33.73 \\
 & APAN & 22.90 & 0.9966 & 0.9966\,{\footnotesize $(0.00\%\downarrow)$} & 0.9970 & 0.9970\,{\footnotesize $(0.00\%\uparrow)$} & 11.20 \\
 & JODIE & 10.59 & 0.9975 & 0.9975\,{\footnotesize $(0.00\%\uparrow)$} & 0.9978 & 0.9978\,{\footnotesize $(0.00\%\downarrow)$} & 3.05 \\
\midrule
\multirow{3}{*}{WIKI-TALK} & TGN & 24.24 & 0.9555 & 0.9563\,{\footnotesize $(0.08\%\uparrow)$} & 0.9505 & 0.9512\,{\footnotesize $(0.08\%\uparrow)$} & 4.60 \\
 & APAN & 4.32 & 0.9363 & 0.9373\,{\footnotesize $(0.11\%\uparrow)$} & 0.9265 & 0.9287\,{\footnotesize $(0.23\%\uparrow)$} & 2.18 \\
 & JODIE & 4.35 & 0.9555 & 0.9544\,{\footnotesize $(0.12\%\downarrow)$} & 0.9493 & 0.9479\,{\footnotesize $(0.15\%\downarrow)$} & 25.62 \\
\midrule
\multirow{3}{*}{SX-FULL} & TGN & 78.66 & 0.9585 & 0.9578\,{\footnotesize $(0.07\%\downarrow)$} & 0.9539 & 0.9530\,{\footnotesize $(0.09\%\downarrow)$} & 40.88 \\
 & APAN & 46.31 & 0.9413 & 0.9418\,{\footnotesize $(0.05\%\uparrow)$} & 0.9385 & 0.9410\,{\footnotesize $(0.27\%\uparrow)$} & 27.47 \\
 & JODIE & 43.32 & 0.9697 & 0.9696\,{\footnotesize $(0.01\%\downarrow)$} & 0.9618 & 0.9616\,{\footnotesize $(0.02\%\downarrow)$} & 15.97 \\
\midrule
\multirow{3}{*}{GDELT} & TGN & 93.13 & 0.9876 & 0.9818\,{\footnotesize $(0.59\%\downarrow)$} & 0.9891 & 0.9846\,{\footnotesize $(0.45\%\downarrow)$} & 52.56 \\
 & APAN & 10.58 & 0.9787 & 0.9761\,{\footnotesize $(0.26\%\downarrow)$} & 0.9836 & 0.9820\,{\footnotesize $(0.16\%\downarrow)$} & 5.39 \\
 & JODIE & 97.11 & 0.9795 & 0.9823\,{\footnotesize $(0.29\%\uparrow)$} & 0.9829 & 0.9853\,{\footnotesize $(0.25\%\uparrow)$} & 28.95 \\
\bottomrule
\end{tabular*}
\endgroup

%% file: tables/eval_components.tex
\begingroup
\small
\setlength{\tabcolsep}{1.5pt}
\begin{tabular*}{\linewidth}{@{\extracolsep{\fill}}llcccccc@{}}
\toprule
& & \multicolumn{3}{c}{WIKI} & \multicolumn{3}{c}{REDDIT} \\
\cmidrule(lr){3-5}\cmidrule(lr){6-8}
Model & Filters & Input($\downarrow$\%) & AUC & Time($\downarrow$\%) & Input($\downarrow$\%)& AUC & Time($\downarrow$\%) \\
\midrule
\multirow{4}{*}{TGN} & Base & 0.00 & 0.9949 & 0.00 & 0.00 & 0.9984 & 0.00 \\
 & Root only & 15.00 & 0.9950\,{\footnotesize $(0.01\%\uparrow)$} & 8.15 & 50.42 & 0.9982\,{\footnotesize $(0.02\%\downarrow)$} & 36.90 \\
 & Support only & 11.07 & 0.9946\,{\footnotesize $(0.02\%\downarrow)$} & 4.16 & 56.02 & 0.9983\,{\footnotesize $(0.01\%\downarrow)$} & 37.69 \\
 & Both (\OurMethod{}) & 16.74 & 0.9947\,{\footnotesize $(0.02\%\downarrow)$} & 9.08 & 67.68 & 0.9981\,{\footnotesize $(0.03\%\downarrow)$} & 48.32 \\
\midrule
\multirow{4}{*}{APAN} & Base & 0.00 & 0.9819 & 0.00 & 0.00 & 0.9970 & 0.00 \\
 & Root only & 5.45 & 0.9794\,{\footnotesize $(0.26\%\downarrow)$} & 1.39 & 18.61 & 0.9970\,{\footnotesize $(0.00\%\downarrow)$} & 9.33 \\
 & Support only & 4.22 & 0.9821\,{\footnotesize $(0.02\%\uparrow)$} & 0.18 & 11.93 & 0.9973\,{\footnotesize $(0.02\%\uparrow)$} & -0.72 \\
 & Both (\OurMethod{}) & 6.55 & 0.9771\,{\footnotesize $(0.49\%\downarrow)$} & 0.56 & 22.90 & 0.9973\,{\footnotesize $(0.02\%\uparrow)$} & 9.60 \\
\bottomrule
\end{tabular*}
\endgroup

%% file: tables/eval_sensitivity_main.tex
\begingroup
\small
\setlength{\tabcolsep}{2pt}
\begin{tabular*}{\linewidth}{@{\extracolsep{\fill}}llrrrrrr@{}}
\toprule
& & \multicolumn{2}{c}{$\tau=0.8$} & \multicolumn{2}{c}{$\tau=0.9$} & \multicolumn{2}{c}{$\tau=0.95$} \\
\cmidrule(lr){3-4}\cmidrule(lr){5-6}\cmidrule(lr){7-8}
Dataset & Model & Input($\downarrow$\%) & $\Delta$AUC (\%) & Input($\downarrow$\%)& $\Delta$AUC (\%) & Input($\downarrow$\%) & $\Delta$AUC (\%) \\
\midrule
\multirow{2}{*}{WIKI} & TGN & 13.39 & +0.07 & 13.82 & +0.04 & 14.10 & +0.05 \\
 & APAN & 6.34 & -0.28 & 6.80 & +0.10 & 6.33 & -0.37 \\
\midrule
\multirow{2}{*}{REDDIT} & TGN & 52.98 & -0.01 & 51.62 & -0.01 & 52.78 & -0.01 \\
 & APAN & 22.97 & +0.02 & 27.53 & +0.00 & 25.41 & +0.00 \\
\bottomrule
\end{tabular*}
\endgroup

%% file: sections/conclusion.tex
In this work, we examine whether all temporal events are necessary for TGNN training. We propose \OurMethod{}, which uses node memory changes to identify redundancy in prediction targets and sampled historical context. Its root-event and support-event filters reduce inputs before they trigger downstream processing, avoiding work in the dependency-constrained training pipeline. Evaluations across three TGNN models and five datasets demonstrate substantial input and training-time reductions, with validation quality remaining close to the baseline. The gains on large event streams highlight input selection as a practical approach to improving TGNN training efficiency as temporal graphs grow.

%% file: sections/appendix.tex

\section{Experimental Setup}
\label{app:setup}

\subsection{Datasets, models, and training frameworks}
We evaluate three memory-based TGNNs, TGN~\citep{rossi2020temporal}, APAN~\citep{wang2021apan}, and JODIE~\citep{kumar2019predicting}, on five temporal graphs. Table~\ref{tab:datasets} lists their node counts, event counts, and interaction-feature dimensions. We abbreviate STACK-OVERFLOW as SX-FULL throughout the evaluation. WIKI-TALK and SX-FULL use constructed 32-dimensional interaction features. Together, the datasets span approximately 157 thousand to 191 million events, allowing us to examine input selection across substantially different training workloads.

\begin{table}[htbp]
  \centering\small
  \caption{Dataset statistics and interaction-feature dimensions. SX-FULL denotes STACK-OVERFLOW. Events are timestamped interactions in the original graph.}
  \label{tab:datasets}
  \begin{tabular*}{\linewidth}{@{\extracolsep{\fill}}lrrr@{}}
\toprule
Dataset & Nodes & Events & Feature dim. \\
\midrule
WIKI~\citep{kumar2019predicting} & 9,227 & 157,474 & 172 \\
REDDIT~\citep{kumar2019predicting} & 11,000 & 672,447 & 172 \\
WIKI-TALK~\citep{paranjape2017motifs} & 1,140,149 & 7,833,140 & 32 \\
SX-FULL~\citep{paranjape2017motifs} & 2,601,977 & 63,497,050 & 32 \\
GDELT~\citep{leetaru2013gdelt,zhou2022tgl} & 16,682 & 191,290,882 & 186 \\
\bottomrule
\end{tabular*}
\end{table}

Our primary comparison adds \OurMethod{} to Cascade~\citep{dai2025cascade}; the additional framework comparison adds it to TGL~\citep{zhou2022tgl}. Base denotes the corresponding model trained without \OurMethod{}. Each comparison retains the underlying framework's optimizer and temporal-sampling configuration. TGN and APAN use both root-event and support-event filtering. JODIE does not sample historical neighbors, so it uses only the root-event filter. Experiments run on an AMD EPYC 7742 64-core CPU and an NVIDIA A100 GPU with 40\,GB of memory, using CUDA 11.6.

\subsection{Training configurations}
Table~\ref{tab:eval-main-configs} summarizes the epoch budgets for the main Cascade comparisons. A node is classified as stable when the cosine similarity between its memories before and after its latest update exceeds $\tau$. We use $\tau=0.9$ unless otherwise specified; the main comparisons use $\tau=0.95$ for TGN on GDELT and for JODIE on SX-FULL. The epoch budgets range from five or six epochs on GDELT to 100 epochs on WIKI and REDDIT.

\begin{table}[htbp]
  \centering\small
  \caption{Epoch budgets for the main Cascade comparisons. Base and \OurMethod{} use the same epoch budget within each model--dataset comparison.}
  \label{tab:eval-main-configs}
  \begin{tabular*}{\linewidth}{@{\extracolsep{\fill}}llrrrrr@{}}
\toprule
Model & Setting & WIKI & REDDIT & WIKI-TALK & SX-FULL & GDELT \\
\midrule
TGN & Epochs & 100 & 100 & 30 & 20 & 5 \\
\midrule
APAN & Epochs & 100 & 100 & 30 & 20 & 5 \\
\midrule
JODIE & Epochs & 100 & 100 & 30 & 20 & 6 \\
\bottomrule
\end{tabular*}
\end{table}

\paragraph{Component and threshold studies.}
\label{app:study-settings}
The component study evaluates TGN and APAN on WIKI and REDDIT for 100 epochs at $\tau=0.9$. Base disables both filters; Root only and Support only enable the indicated filter; Both enables both filters. The sensitivity study evaluates Both for 100 epochs at $\tau\in\{0.8,0.9,0.95\}$. Both studies use a maximum batch size of 8,000 and one history window. Each model--dataset comparison uses a common unfiltered baseline for its variants and holds the remaining settings fixed.

\subsection{Evaluation protocol and measurements}
Events follow the framework's chronological training, validation, and test splits. Evaluation samples one negative interaction per observed interaction and averages AP and AUC over evaluation batches. The checkpoint is selected by validation AP after the first three epochs, and AUC is taken from the same checkpoint. Evaluation processes events chronologically and updates node memories with filtering disabled. All AP and AUC values reported in this paper are validation measurements.

\paragraph{Training inputs.}
We collectively refer to root events and support events as \emph{training inputs}. Root events provide prediction targets, while support events provide historical context for those targets. For a given input stream, reduction is
\begin{equation}
 R_{\mathrm{input}}=100\left(1-\frac{N_{\mathrm{RAFT}}}{N_{\mathrm{base}}}\right)\%,
 \label{eq:prune}
\end{equation}
where $N$ counts root events for JODIE and sampled historical-event occurrences for TGN and APAN. Repeated sampling of the same interaction contributes multiple occurrences. Root filtering reduces the number of prediction targets and avoids their sampling requests, while support filtering reduces the context returned for retained targets. Main-table averages assign equal weight to the 15 model--dataset configurations.

\paragraph{Prediction quality.}
AP and AUC quantify validation prediction quality, with higher values indicating better predictions. For either metric $q$, we report its relative change as
\begin{equation}
 \Delta q=100\frac{q_{\mathrm{RAFT}}-q_{\mathrm{base}}}{q_{\mathrm{base}}}\%.
 \label{eq:appendix-quality}
\end{equation}
Positive values indicate an increase; negative values indicate a decrease. Mean absolute relative changes average $|\Delta q|$ across the stated comparisons. All percentages are computed before rounding the displayed scores.

\paragraph{Training time.}
Training-time reduction is $100(1-T_{\mathrm{RAFT}}/T_{\mathrm{base}})\%$, using the same recorded timing field for both configurations. Framework comparisons use recorded training time; stage profiles report mean times per epoch. Appendix~\ref{app:timer-definitions} defines these timing scopes. All reported averages are arithmetic means over the specified configurations.

\newpage
\section{Additional Motivation Results}
\label{app:extra-motivation}
We extend the motivation in two directions: whether stable memories consistently identify root events with smaller prediction losses, and whether historical inputs associated with stable memories can be removed while retaining the observed loss-reduction trend.

\subsection{Stable root events across models and datasets}
\label{app:root}
We classify a root event as stable when both its source and destination nodes have memory cosine similarity above $0.9$ between their pre-update and post-update states. Table~\ref{tab:extra-root} compares stable and unstable events across all three models and five datasets. It reports their mean positive-event prediction losses, together with stable events' shares of observed event occurrences and aggregate positive-event loss.

\begin{table}[htbp]
  \centering\small
  \caption{Root-event motivation results across three models and five datasets. Stable and Unstable report mean positive-event prediction loss. Events and Loss report stable events' shares of observed root-event occurrences and aggregate positive-event loss, respectively.}
  \label{tab:extra-root}
  \begin{tabular*}{\linewidth}{@{\extracolsep{\fill}}llrrrr@{}}
\toprule
& & \multicolumn{2}{c}{Mean positive-event loss} & \multicolumn{2}{c}{Stable-event share (\%)} \\
\cmidrule(lr){3-4}\cmidrule(lr){5-6}
Model & Dataset & Stable & Unstable & Events & Loss \\
\midrule
TGN & WIKI & 0.0668 & 0.2767 & 12.98 & 3.48 \\
 & REDDIT & 0.0302 & 0.0939 & 36.16 & 15.40 \\
 & WIKI-TALK & 0.0099 & 0.3793 & 17.69 & 0.56 \\
 & SX-FULL & 0.0414 & 0.5897 & 63.59 & 10.92 \\
 & GDELT & 0.1134 & 1.2094 & 98.91 & 89.47 \\
\midrule
APAN & WIKI & 0.1081 & 0.3630 & 2.27 & 0.69 \\
 & REDDIT & 0.0617 & 0.1203 & 10.58 & 5.72 \\
 & WIKI-TALK & 0.0156 & 0.3987 & 1.55 & 0.06 \\
 & SX-FULL & 0.0250 & 0.3107 & 16.08 & 1.52 \\
 & GDELT & 0.0924 & 0.1482 & 10.46 & 6.79 \\
\midrule
JODIE & WIKI & 0.0608 & 0.3394 & 2.02 & 0.37 \\
 & REDDIT & 0.0291 & 0.1005 & 10.59 & 3.32 \\
 & WIKI-TALK & 0.0074 & 0.3602 & 4.35 & 0.09 \\
 & SX-FULL & 0.0346 & 0.4253 & 43.32 & 5.85 \\
 & GDELT & 0.1162 & 0.9146 & 97.11 & 81.00 \\
\bottomrule
\end{tabular*}
\end{table}

Stable events have lower mean loss in all 15 settings. Their share of aggregate loss is consequently smaller than their share of event occurrences. On SX-FULL, for example, stable events account for 63.59\% of TGN's observed roots but only 10.92\% of its positive-event loss. The corresponding shares are 16.08\% versus 1.52\% for APAN and 43.32\% versus 5.85\% for JODIE. On WIKI-TALK, stable events contribute less than 0.6\% of positive-event loss for each model. These results extend the main observation beyond a single architecture: memory stability identifies a population of targets that the model predicts with comparatively small error, motivating their selection before they trigger further training work.

\subsection{Memory changes and prediction loss}
The stable/unstable comparison groups events using a threshold. We further examine the relationship across the range of memory changes using Spearman rank correlation, which measures whether larger changes tend to accompany larger losses. For a root event, the change score is one minus the average cosine similarity of its source and destination memories. For sampled support, the score is one minus the average memory cosine similarity over the sampled support nodes. Each score is compared with the corresponding target's positive-event prediction loss.

\begin{table}[htbp]
  \centering\small
  \caption{Spearman correlations between memory-change scores and positive-event prediction loss. Root scores average source and destination memory changes; support scores average changes over sampled support nodes. Positive values indicate that larger memory changes tend to accompany larger losses.}
  \label{tab:eval-correlation}
  \begin{tabular*}{\linewidth}{@{\extracolsep{\fill}}lrrrr@{}}
\toprule
& \multicolumn{3}{c}{Root memory change} & Support memory change \\
\cmidrule(lr){2-4}\cmidrule(lr){5-5}
Dataset & TGN & APAN & JODIE & TGN \\
\midrule
WIKI & 0.445 & 0.703 & 0.625 & 0.405 \\
REDDIT & 0.287 & 0.280 & 0.289 & 0.158 \\
WIKI-TALK & 0.666 & 0.724 & 0.679 & 0.458 \\
SX-FULL & 0.402 & 0.488 & 0.474 & 0.019 \\
\bottomrule
\end{tabular*}
\end{table}

Table~\ref{tab:eval-correlation} shows positive root-score correlations in all 12 recorded settings, ranging from 0.280 to 0.724. This relationship complements the threshold-based results and supports memory change as a signal for root-event selection. Support-score correlations vary more, from 0.019 to 0.458 for TGN. Because historical inputs contribute jointly to a target's prediction, we next examine their redundancy directly by removing selected support events and observing training loss.

\subsection{Support-event filtering on WIKI-TALK}
\label{app:convergence}
\begin{figure}[htbp]
  \centering
  \includegraphics[width=\linewidth]{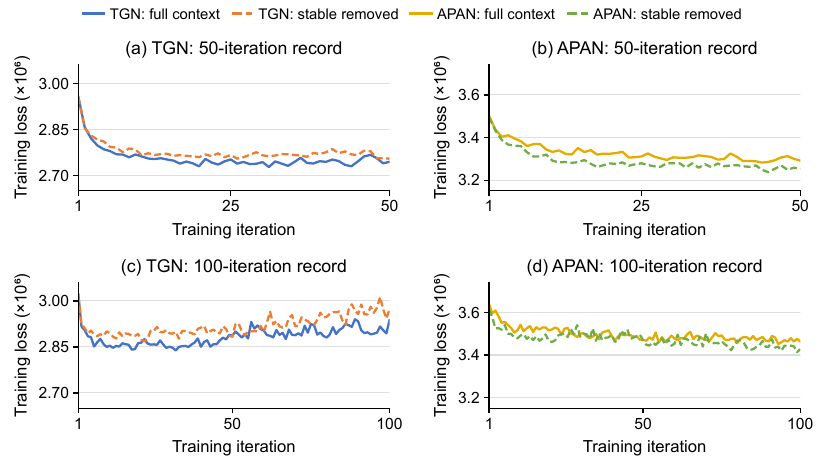}
  \caption{Training-loss curves with and without stable support events on WIKI-TALK. Solid curves use full sampled history; dashed curves remove support events whose source nodes are stable. The 50- and 100-iteration panels show separate studies.}
  \label{fig:extra-support}
  \label{fig:convergence}
\end{figure}

The main motivation compares training with full sampled history and with stable support events removed on WIKI and REDDIT. Figure~\ref{fig:extra-support} extends this comparison to WIKI-TALK. In both configurations, the prediction targets are retained; the filtered configuration omits historical events whose source nodes are stable. The figure contains separate 50- and 100-iteration studies for TGN and APAN.

In the 50-iteration study, TGN's paired curves remain close, and APAN follows a similar declining trajectory with lower loss under filtering. The 100-iteration study again shows close APAN trajectories, while TGN's filtered loss remains higher during much of the later period. Together with the main motivation, these results show that part of the sampled history can be removed while maintaining similar loss reduction in several settings. The differing TGN trajectories also motivate evaluating input savings together with prediction quality when selecting the filtering strength.

\newpage
\section{Efficiency Analysis}
\label{app:timing}
Input filtering reduces the events that reach subsequent training stages. This section examines where the resulting time savings arise, how much stability maintenance costs, and how the approach complements different training frameworks.

\subsection{Training stages and timing definitions}
\label{app:timer-definitions}
Table~\ref{tab:timer-definitions} defines the recorded timers. Sample covers target preparation and historical sampling. Prep covers preparing model inputs and maintaining node states, including state writes after model execution. Model covers prediction and parameter updates. Total encloses these batch-processing operations and accumulates their elapsed time over training.

\begin{table}[htbp]
  \centering\small
  \caption{Training-stage and stability-maintenance timers. Lookup and Update may lie within enclosing stage timers and are reported separately rather than added to Total.}
  \label{tab:timer-definitions}
  \begin{tabular*}{\linewidth}{@{}p{.14\linewidth}p{\dimexpr.86\linewidth-2\tabcolsep\relax}@{}}
  \toprule
  Timer & Measured operations \\
  \midrule
  Sample & Preparing root and negative targets, including applicable root filtering, and sampling historical neighbors. \\
  \addlinespace[3pt]
  Prep & Building computation blocks; gathering features, memories, and messages; transferring required data; and committing updated memories and messages after model execution. \\
  \addlinespace[3pt]
  Model & Forward prediction, loss computation, backpropagation, and optimizer updates, including memory-update computation inside the model. \\
  \addlinespace[3pt]
  Total & The enclosing interval from batch input processing through model execution and post-update state maintenance. \\
  \addlinespace[3pt]
  Lookup & Reading stored node-stability flags for input selection. \\
  \addlinespace[3pt]
  Update & Computing memory cosine similarities and updating the stability flags during node-state maintenance. \\
  \bottomrule
  \end{tabular*}
\end{table}

\subsection{Preparation savings and stability-maintenance cost}
Preparation is the largest measured training stage in all four profiling settings, accounting for 75--82\% of baseline Total. Table~\ref{tab:eval-profile-raw} compares the stage times for TGN and APAN on REDDIT and SX-FULL. The measurements are mean times per epoch over ten profiled epochs on REDDIT and five on SX-FULL.

\begin{table}[htbp]
  \centering\small
  \caption{Mean training-stage times per epoch (seconds) for Base and \OurMethod{}. REDDIT profiles average ten epochs; SX-FULL profiles average five. Total is the enclosing batch-processing timer defined in Table~\ref{tab:timer-definitions}.}
  \label{tab:eval-profile-raw}
  \setlength{\tabcolsep}{3pt}
  \begin{tabular*}{\linewidth}{@{\extracolsep{\fill}}lllrrrr@{}}
\toprule
Model & Dataset & Method & Sample (s) & Prep (s) & Model (s) & Total (s) \\
\midrule
TGN & REDDIT & Base & 0.441 & 8.864 & 1.868 & 11.175 \\
 &  & \OurMethod{} & 0.410 & 6.095 & 1.805 & 8.308 \\
\midrule
TGN & SX-FULL & Base & 22.154 & 241.630 & 55.350 & 319.146 \\
 &  & \OurMethod{} & 18.016 & 128.142 & 42.008 & 188.176 \\
\midrule
APAN & REDDIT & Base & 0.442 & 6.611 & 1.707 & 8.761 \\
 &  & \OurMethod{} & 0.559 & 5.669 & 1.652 & 7.884 \\
\midrule
APAN & SX-FULL & Base & 17.674 & 184.968 & 24.132 & 226.788 \\
 &  & \OurMethod{} & 17.976 & 156.986 & 25.648 & 200.624 \\
\bottomrule
\end{tabular*}
\end{table}

\OurMethod{} reduces Prep in all four settings, with savings of 14.25--46.97\%. The largest reduction occurs for TGN on SX-FULL, where Prep decreases from 241.630 to 128.142 seconds per epoch. Total decreases from 319.146 to 188.176 seconds, a 41.04\% reduction in the enclosing training interval. These measurements connect input selection to the work it avoids: fewer event inputs require less feature and state handling, reducing the dominant preparation cost.

\begin{table}[htbp]
  \centering\small
  \caption{Stability-maintenance cost and preparation-time reduction. Lookup and Update are mean times per epoch in seconds. Stability cost is their sum as a percentage of Base Total; Prep reduction uses Base Prep as its denominator. Percentages use unrounded timer values.}
  \label{tab:eval-profile}
  \setlength{\tabcolsep}{3pt}
  \begin{tabular*}{\linewidth}{@{\extracolsep{\fill}}llrrrr@{}}
\toprule
Model & Dataset & Lookup (s) & Update (s) & \shortstack{Stability cost\\(\% of Base Total)} & \shortstack{Prep reduction\\(\%)} \\
\midrule
TGN & REDDIT & 0.0067 & 0.1621 & 1.51 & 31.24 \\
TGN & SX-FULL & 0.1929 & 6.9066 & 2.22 & 46.97 \\
APAN & REDDIT & 0.0121 & 0.2564 & 3.07 & 14.25 \\
APAN & SX-FULL & 0.2102 & 6.8293 & 3.10 & 15.13 \\
\bottomrule
\end{tabular*}
\end{table}

The filtering signal requires comparatively little maintenance. As shown in Table~\ref{tab:eval-profile}, Lookup and Update together account for 1.51--3.10\% of baseline Total. For TGN on SX-FULL, their combined cost is approximately 7.10 seconds per epoch, compared with 113.49 seconds of preparation time saved. The stage and stability measurements therefore show that input selection can avoid substantial preparation work with a relatively small cost for maintaining memory-stability information.

\subsection{Integration with TGL}
The time savings also extend to TGL~\citep{zhou2022tgl}. Table~\ref{tab:eval-tgl} pairs validation quality with recorded net-time reduction across all 15 model--dataset comparisons. \OurMethod{} reduces training time by 12.42\% on average, complementing TGL's data preparation and execution optimizations. TGN on GDELT and JODIE on SX-FULL obtain reductions of 29.10\% and 32.82\%, respectively.

\begin{table}[htbp]
  \centering\small
  \caption{Comparison between models trained under TGL (Base) and TGL with \OurMethod{}. AP and AUC are validation scores. Time reduction is the percentage decrease in recorded training time relative to the corresponding Base configuration.}
  \label{tab:eval-tgl}
  \begingroup
\small
\setlength{\tabcolsep}{2pt}
\begin{tabular*}{\linewidth}{@{\extracolsep{\fill}}llrrrrr@{}}
\toprule
\multirow{2}{*}{Dataset} & \multirow{2}{*}{Model} & \multicolumn{2}{c}{Validation AP} & \multicolumn{2}{c}{Validation AUC} & \multirow{2}{*}{\shortstack{Time reduction\\(\%)}} \\
\cmidrule(lr){3-4}\cmidrule(lr){5-6}
& & Base & \OurMethod{} & Base & \OurMethod{} & \\
\midrule
\multirow{3}{*}{WIKI} & TGN & 0.9962 & 0.9952 & 0.9962 & 0.9953 & 10.85 \\
 & APAN & 0.9862 & 0.9852 & 0.9874 & 0.9870 & 6.22 \\
 & JODIE & 0.9823 & 0.9823 & 0.9837 & 0.9845 & 5.33 \\
\midrule
\multirow{3}{*}{REDDIT} & TGN & 0.9984 & 0.9985 & 0.9986 & 0.9986 & 16.33 \\
 & APAN & 0.9975 & 0.9975 & 0.9978 & 0.9978 & 8.44 \\
 & JODIE & 0.9976 & 0.9974 & 0.9978 & 0.9977 & 4.00 \\
\midrule
\multirow{3}{*}{WIKI-TALK} & TGN & 0.9591 & 0.9597 & 0.9543 & 0.9552 & 5.00 \\
 & APAN & 0.9447 & 0.9438 & 0.9367 & 0.9352 & 2.72 \\
 & JODIE & 0.9560 & 0.9575 & 0.9496 & 0.9511 & 27.72 \\
\midrule
\multirow{3}{*}{SX-FULL} & TGN & 0.9722 & 0.9713 & 0.9664 & 0.9655 & 12.24 \\
 & APAN & 0.9461 & 0.9445 & 0.9435 & 0.9421 & 14.37 \\
 & JODIE & 0.9730 & 0.9721 & 0.9657 & 0.9649 & 32.82 \\
\midrule
\multirow{3}{*}{GDELT} & TGN & 0.9854 & 0.9848 & 0.9874 & 0.9868 & 29.10 \\
 & APAN & 0.9817 & 0.9787 & 0.9852 & 0.9835 & 8.21 \\
 & JODIE & 0.9865 & 0.9849 & 0.9882 & 0.9869 & 2.99 \\
\bottomrule
\end{tabular*}
\endgroup
\end{table}